\documentclass[letterpaper]{article} 
\usepackage{aaai2026}  
\makeatletter
\gdef\copyright@on{F}  
\gdef\copyright@text{} 
\makeatother
\usepackage{times}  
\usepackage{helvet}  
\usepackage{courier}  
\usepackage[hyphens]{url}  
\usepackage{graphicx} 
\usepackage{natbib}  
\usepackage{caption} 
\usepackage{algorithm}
\usepackage{algorithmic}
\usepackage{booktabs}
\usepackage{amsmath}
\usepackage{comment}
\usepackage{amsfonts}
\usepackage{xcolor}

\usepackage{newfloat}
\usepackage{listings}

\DeclareCaptionStyle{ruled}{labelfont=normalfont,labelsep=colon,strut=off} 
\floatstyle{ruled}
\newfloat{listing}{tb}{lst}{}
\floatname{listing}{Listing}
\title{Improving Factuality for Dialogue Response Generation via Graph-Based Augmentation}
\author{
    Xiangyan Chen\textsuperscript{\rm 1}, Yujian Gan\textsuperscript{\rm 1}, Yimeng Gu\textsuperscript{\rm 1}, and Matthew Purver\textsuperscript{\rm 1, 2}\
}
\affiliations{
    \textsuperscript{\rm 1}Queen Mary University of London, UK\\
    \textsuperscript{\rm 2}Institut Jožef Stefan, Slovenia\\

    {\{xiangyan.chen, y.gan, yimeng.gu, m.purver\}@qmul.ac.uk}
}

\usepackage{bibentry}

\begin{document}

\maketitle

\begin{abstract}
Large Language Models (LLMs) succeed in many natural language processing tasks. However, their tendency to hallucinate --- generate plausible but inconsistent or factually incorrect text --- can cause significant problems in certain tasks, including response generation in dialogue. 
To mitigate this issue, we propose two novel graph knowledge-augmented frameworks, Dialogue Response Generation via Textualised Graphs (TG-DRG) and Graph-Aware Dialogue Response Generation (GA-DRG), which combine reasoning-guided dialogue reformulation, dialogue sense knowledge selection, and graph-enhanced response generation to improve the factuality of dialogue responses. To evaluate the factuality of generated responses, we propose a dialogue fact score that addresses the limitations of existing fact-score methods in dialogue settings, providing a more reliable assessment of factual consistency. We evaluate our methods using different baselines on the OpendialKG and HybriDialogue datasets. 
Our methods noticeably improve factuality compared to other graph knowledge-augmentation baselines, including the state-of-the-art G-retriever, achieving improvements of 3.47\% on OpendialKG and 3.12\% on HybriDialogue in terms of dialogue fact score. The code will be released on GitHub\footnote{\url{https://github.com/XiangyanChen/DRG}}.
\end{abstract}


\section{Introduction}

Large Language Models (LLMs) have been shown to perform powerfully on Natural Language Processing (NLP) tasks. Despite their general superiority, LLMs will generate some plausible but fact-inconsistent text, namely hallucination. 
Flawed pre-training data, model bias and randomness in inference are the factors contributing to hallucinations \cite{zhang2023siren, huang2023survey}. In dialogue response generation, generating incorrect responses will mislead people and have a negative impact on society.

A number of approaches have been proposed to enhance the factuality of LLMs. Among them, graph knowledge-augmented methods aim to improve factual accuracy by incorporating knowledge graphs. This has been effective in tasks such as Question Answering (QA) \cite{baek2023knowledge, wu2023retrieve, he2024g}.

\begin{figure}[t]
\centering
  \includegraphics[width=1\linewidth]{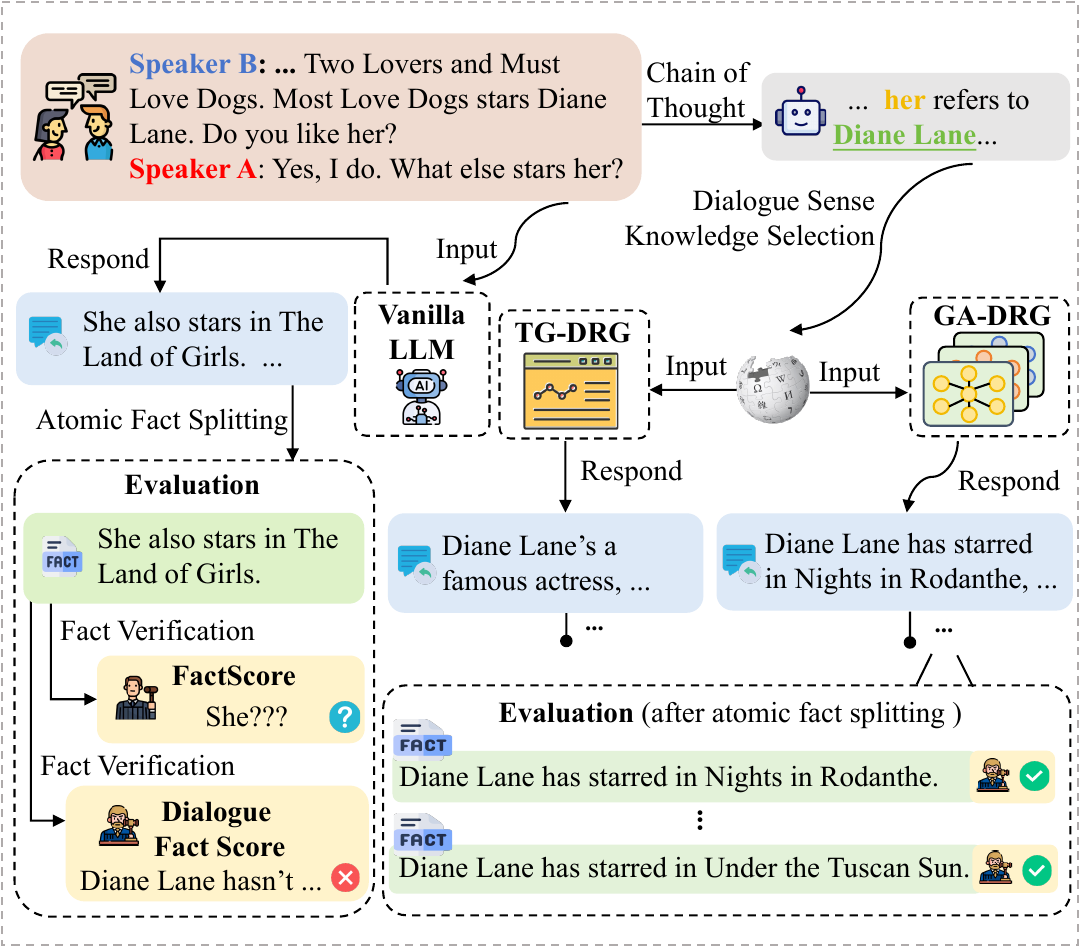}
  \caption {An example illustrates how our frameworks TG-DRG and GA-DRG, and dialogue fact score work.}
  \label{figure:dialogue}
\end{figure}

Recently,
\citet{baek2023knowledge} utilised sentence embedders to encode knowledge and determine the match between a query and knowledge based on similarity. \citet{he2024g} proposed a graph-based method called G-retriever, which employs the Prize-Collecting Steiner Tree (PCST) to help construct a subgraph. However, applying these methods to dialogue response generation is challenging due to their lack of consideration for dialogue context. Dialogues frequently involve intricate coreference structures, which hinder the LLMs’ comprehension, ultimately compromising the factual consistency and quality of the generated responses. In addition, evaluating factuality for dialogue responses is a challenge. F1-score relying on entities overlapping has limitations on open-domain tasks. More advanced FactScore \cite{min2023factscore} evaluates facuality by splitting long-text into smaller atomic facts, but it does not consider dialogue context and situations where the knowledge source is unavailable, making it difficult to evaluate dialogue responses.

To overcome the limitation on generating factually correct responses, we propose two frameworks: (1) \emph{Dialogue Response Generation via Textualised Graphs (TG-DRG)}, where generation requires no training; and (2) \emph{Graph Aware Dialogue Response Generation (GA-DRG)}, which explicitly models node relationships, designed to improve the factual accuracy of dialogue generation. Both frameworks are built around reasoning-guided dialogue reformulation and dialogue sense knowledge selection. The latter helps select the most valuable knowledge based on the dialogue. We use an example shown in Figure~\ref{figure:dialogue} to illustrate how our frameworks work. The response to the dialogue is incorrect due to the pronouns, e.g. ``her''. After applying Chain of Thought (CoT), the pronoun is solved, benefiting further knowledge-augmented dialogue response generation.

To address limitations in evaluating factual accuracy, we propose the \emph{Dialogue Fact Score}, as illustrated in Figure~\ref{figure:dialogue}. In this case, the FactScore can not work without a dialogue context. Our metric accounts for both the dialogue context and scenarios in which the knowledge source is partially or entirely unavailable. Additionally, we propose \emph{Not Enough Information Proportion (NEIP)} to evaluate dialogue factuality comprehensively. It is a metric that quantifies the proportion of atomic facts in a response that cannot be verified, such as opinions or hallucinated content that can not find any evidence to verify. We assess model-human agreement of dialogue fact score. Annotation results show that our dialogue fact score achieves substantial model-human agreement.

We compare our proposed method not only with standalone LLMs but also with knowledge-enhanced approaches such as KAPING \cite{baek2023knowledge}, the BM25 algorithm, and the State-of-the-Art (SOTA) G-retriever \cite{he2024g}. Experimental results show that our framework consistently outperforms both existing knowledge-enhanced methods and the current best-performing G-retriever.  

Our contributions to this work can be listed as:
\begin{enumerate}

\item We propose the dialogue fact score to evaluate the factuality of dialogue systems automatically. The evaluation is fine-grained, enabling a more reliable assessment of factual consistency in dialogue systems.

\item We propose an unsupervised framework, TG-DRG, designed to enhance the factual accuracy of responses without generation training. This framework adopts a textualised graph to generate responses built around reasoning-guided dialogue reformulation and dialogue sense knowledge selection. We validate TG-DRG against various comparable baselines; the experimental results show that TG-DRG can improve factuality noticeably.

\item We propose a supervised framework, GA-DRG. Different from TG-DRG, GA-DRG utilises a graph-based neural network to capture the node relationships, making it more efficient in improving factuality when generating dialogue responses. The experimental results show that GA-DRG outperforms the SOTA G-retriever in dialogue fact score with an improvement of 3.47\% on OpendialKG and 3.12\% on HybriDialogue. 

\end{enumerate}

\begin{figure*}[ht]
\centering
  \includegraphics[width=1\linewidth]{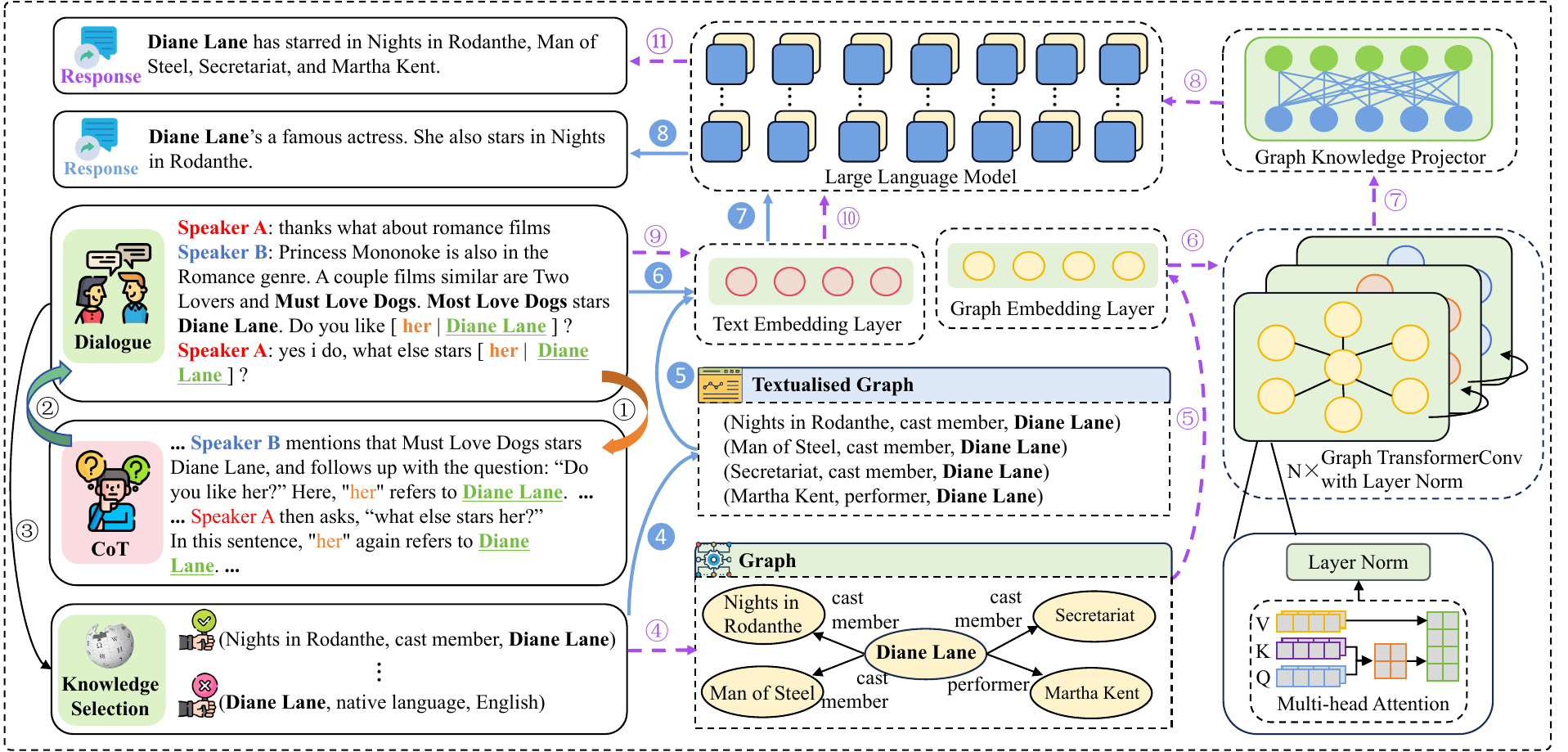}
  \caption{The workflow of our proposed frameworks, TG-DRG and GA-DRG, begins with the dialogue. A CoT process first reformulates the dialogue. Based on the refined dialogue, the most valuable knowledge triples are then selected by the dialogue sense knowledge selection module. Starting from Step 4, the workflow diverges into two distinct paths for dialogue response generation: the blue solid arrows represent the TG-DRG path, while the purple dashed arrows indicate the GA-DRG path.}
  \label{figure:framework}
\end{figure*}

\section{Related Work}
\subsection{LLM Hallucinations and Mitigation Methods}

LLMs are trending in NLP. However, hallucination exists in these LLMs widely, which is the phenomenon that language models generate coherent but fact-inconsistent text. 

Several knowledge-based methods have been proposed to mitigate fact-conflicting hallucinations \cite{agrawal2023can}, which generally fall into three categories: knowledge-aware inference, training, and validation. For instance, \citet{baek2023knowledge} introduced KAPING, a prompt-based framework for QA that retrieves knowledge triples based on embedding similarity. Building upon retrieval techniques, \citet{sen2023knowledge} combined Knowledge Graph (KG) retrieval with language model reasoning to enhance performance on complex questions. Similarly, CRAG \cite{yan2024corrective} employs an evaluator to assess generation quality and, if necessary, refines outputs via web search, showing significant improvements across various generation tasks. Another notable approach is Self-RAG \cite{asai2023self}, which retrieves relevant knowledge, generates an initial answer, and then refines it through self-critique, achieving strong results in multiple question answering benchmarks. In a related development, \citet{he2024g} proposed G-Retriever, which integrates a graph-based encoder with the PCST algorithm to encode retrieved KG effectively, standing out among graph-enhanced knowledge retrieval methods.

However, applying QA-based approaches to multi-turn dialogue generation is limited by their lack of coreference modelling, which is essential for capturing discourse-level dependencies and ensuring factual consistency.

\subsection{Evaluation of Factuality}
Dialogue response generation not only needs to consider the dialogue context when generating responses but also needs to be evaluated based on the factuality and quality, rather than just the accuracy of answers of QA. Previous dialogue factuality evaluation mainly focuses on humans \cite{ni2023multi, li2022eliciting, yu2022xdai}, which is inefficient.

Automatic evaluating factuality is challenging, however; current methods are either \emph{reference-based} or \emph{reference-free}.

Several QA datasets provide factual references in the form of entities; reference-based evaluation can then be based on entity matching. Other NLP datasets offer text references, so an alternative reference-based method involves matching the extracted entities between the generated text and these references \cite{nan2021entity}. 
Instead, dialogue datasets provide reference responses. To evaluate the factual accuracy of the generated text, we can therefore compare the entities extracted from the generated response with those in the reference response, measured as the F1 score. However, relying solely on this can be inadequate, as the reference may not always provide explicit answers. It is crucial to consider additional metrics to ensure a comprehensive evaluation.

Reference-free methods can be categorised into uncertainty estimation \cite{farquhar2024detecting} and external knowledge-based approaches, with the former often limited by its reliance on the generation model’s own confidence. The FactScore \cite{min2023factscore}, which falls into the latter category, is a metric designed to measure fact consistency in long-form text. The process begins with an LLM breaking down the text into fine-grained sentences called atomic facts. These atomic facts are then verified using both the LLM and external knowledge sources (\citet{min2023factscore} use Wikipedia titles). The FactScore is calculated based on the precision. 

\label{related-work-factscore}
However, the FactScore is not directly suitable for our work for two reasons: First, it evaluates only the generated text without taking the dialogue history into account, which is crucial when assessing the dialogue factuality. Second, it does not account for the possibility that generated responses may lack external knowledge for verification.

\section{Methodology}
To address the limitations of prior works \cite{he2024g, baek2023knowledge}, which often overlook complex pronoun resolution in dialogue settings and struggle to generate responses with high factual accuracy, we propose two novel frameworks:  
(1) \emph{Dialogue Response Generation via Textualised Graphs (TG-DRG)} without training a generation model; and  
(2) \emph{Graph-Aware Dialogue Response Generation (GA-DRG)}, which leverages graph structures to better capture relationships between nodes. Both of them are built around a reasoning-guided coreference resolution and a dialogue sense knowledge selection. TG-DRG includes two variants based on the use of dialogue reformulation:  
\emph{TG-DRG-R} incorporates a reformulation module, while \emph{TG-DRG-NR} bypasses it.  
The overall workflow for our frameworks is illustrated in Figure~\ref{figure:framework}.

In addition, to automatically evaluate factuality in dialogue systems, we propose the \emph{Dialogue Fact Score}, a metric designed to assess factual accuracy specifically within dialogue contexts, thereby addressing the limitations of existing evaluation methods for open-domain dialogue factuality.

\subsection{Task Formulation}

We define the conversation history as a sequence of utterances \(\mathcal{U} = \{ \text{U}_1, \text{U}_2, \dots, \text{U}_{t-1} \}\), where \(\text{U}_i\) denotes the utterance at the \(i^{th}\) turn, either from the user or the system.

In our setup, external knowledge is modeled as a graph \(\mathcal{G} = (\mathcal{V}, \mathcal{E})\), where \(\mathcal{V} = \{v_1, v_2, \dots, v_n\}\) represents the set of nodes, which are entities, and \(\mathcal{E} = \{e_1, e_2, \dots, e_m\}\) represents the set of edges, which are relations. The graph is constructed from an encyclopedia-based knowledge source, with each knowledge fact represented as a triple \((s, p, o)\), where \(s, o \in \mathcal{V}\) denote the head and tail nodes, respectively, and \(p \in \mathcal{E}\) denotes the edge (relation) connecting them.

Our task is to generate a factually correct response \(\text{U}_t\) conditioned on the preceding utterances \(\mathcal{U}\) and the knowledge graph \(\mathcal{G}\).

\subsection{Reasoning-Guided Dialogue and Knowledge Integration}
We introduce a \emph{CoT}-driven approach to dialogue reformulation, which systematically resolves coreferences step-by-step to produce high-quality dialogues. Additionally, we propose a dialogue sense knowledge selection method that constructs dialogue-relevant graphs to enhance factual accuracy. The reformulated dialogue enhances the effectiveness of knowledge selection and further mitigates hallucinations.

\paragraph{Reasoning-Guided Dialogue Reformulation} 
Zero-shot coreference resolution often struggles with resolving complex pronoun references in conversations due to the intricate reasoning involved. To address this challenge, we leverage the reasoning capabilities of LLMs to perform step-by-step resolution for each turn of the dialogue. This iterative approach not only simplifies the understanding of complex coreference relationships but also enhances the model's overall coreference resolution ability. Specifically, we employ the CoT technique to guide the model through incremental reasoning, resulting in more accurate dialogue reformulation and improved factual consistency.

\begin{equation}
\mathcal{U^{'}} = f^{\text{ref}}_{\theta}(p_{\text{reform}}, \mathcal{U}),
\end{equation}
where \( p_{\text{reform}} \) is the prompt guiding the dialogue reformulation process, and \( f^{\text{ref}}_{\theta} \) denotes an LLM that transforms the original dialogue \( \mathcal{U} \) into its reformulated form \( \mathcal{U}' \).

\paragraph{Dialogue Sense Knowledge Selection} 
The external knowledge graph \(\mathcal{G}\) contains noise and dialogue-irrelevant triples, making the LLM difficult to focus on the useful nodes when generating responses. To improve the quality, we propose a dialogue sense selection method relying on an LLM, which helps select the most dialogue-relevant triples and form a dialogue sense graph.

Specifically, we firstly encode the dialogue \(\mathcal{U}\) and the triple \((s, p, o)\), respectively, into vector representations by the encoder \(\psi\) and \(\phi\). The function \(f^{\text{sel}}_{\theta}\) denotes the LLM, which computes the score based on these representations.
The formulation of selection probability is listed as follows:

\begin{equation}
P(y = \text{rel} \mid \mathcal{U'}, (s, p, o)) = \sigma\left(f^{\text{sel}}_{\theta}\left(\psi(\mathcal{U}^{(\cdot)}), \phi(s, p, o)\right)\right),
\end{equation}
where \(\mathcal{U}^{(\cdot)} \in \{ \mathcal{U}, \mathcal{U}^{'}\}\), \(\text{rel}\) denotes relevance. \(\sigma\) is the activation function that maps the output to the probability.

We select the N triples based on the probability from the knowledge graph \(\mathcal{G}\), and the process is defined as follows:

\begin{equation}
\mathcal{G}_{\mathcal{U}^{(\cdot)}} = \text{Top}_N\left(
\mathcal{E},\ 
\left\{ P\left(\text{rel} \mid \mathcal{U}^{(\cdot)},\ (s, p, o)\right) \right\}_{(s,p,o) \in \mathcal{E}}
\right),
\end{equation}
where \(\mathcal{G}_{\mathcal{U}^{(\cdot)}} \in \{\mathcal{G}_{\mathcal{U}},\mathcal{G}_{\mathcal{U}^{'}}\}\) is the dialogue sense graph constructed from the selected triples. 

We select the LLaMA 3 8B model for dialogue reformulation and dialogue sense knowledge selection to achieve a balance between capability and efficiency. To enhance performance and leverage the strengths of advanced models, we fine-tune the LLaMA models using generated samples from GPT-4o, employing a combination of cross-entropy loss and Low-Rank Adaptation (LoRA) techniques \cite{hu2021lora}.

\subsection{Graph Knowledge-Driven Response Generation}
\label{Section:Knowledge-Based-Dialogue-Response-Generation}
The dialogue sense graph facilitates generating factual responses. We propose two graph-based generation methods.

\paragraph{Dialogue Response Generation via Textualised Graphs} Given the dialogue \(\mathcal{U}^{(\cdot)} \in \{ \mathcal{U}, \mathcal{U}^{'}\}\), as well as the dialogue sense graph \(\mathcal{G}_{\mathcal{U}^{(\cdot)}} \in \{\mathcal{G}_{\mathcal{U}},\mathcal{G}_{\mathcal{U}^{'}}\}\), the generation process is formulated as follows:

\begin{equation}
\text{U}_{t}^{(\cdot)} = f^{\text{gen}}_{\theta} \big(p_{\text{gen}},\ \mathcal{U}^{(\cdot)},\ \mathcal{T}_{\mathcal{G}^{(\cdot)}}\big), 
\end{equation}

\noindent where $p_{\text{gen}}$ denotes the prompt for generating the response. \(f^{\text{gen}}_{\theta}\) is the LLM. We refer to the generation \( \text{U}_{t} \) as TG-DRG-NR, and the generation \( \text{U}_{t}^{'} \) as TG-DRG-R. \(\mathcal{T}_{\mathcal{G}^{(\cdot)}}\) denotes a textualised graph, formed of a set of text triples.

\paragraph{Graph Aware Dialogue Response Generation} Textualised graph-based generation can effectively improve factuality without training, while it is unable to capture the connectivity information within a KG and furthermore affects the factuality of the response. To address this limitation, we encode the knowledge graph based on GNNs (Graph Neural Networks) \cite{kipf2016semi}. 

To generate responses with a graph, we aim to encode the dialogue sense graph \(\mathcal{G}_{\mathcal{U}^{'}}\) into a graph-level representation. We firstly encode its nodes \(\mathcal{V}^{'}\):

\begin{equation}
\mathbf{H}^{0}=\text{GraphEmb}(\mathcal{V}') \in \mathbb{R}^{|\mathcal{V}'| \times d},
\end{equation}
where \(d\) is the embedding dimension. The matrix \(\mathbf{H}^0\) consists of node embeddings produced by the Graph Embedding Layer (GraphEmb). To get the higher node representation, we utilise a layer based on multi-head attention \cite{vaswani2017attention}, which is formulated as follows:

\begin{equation}
\mathbf{H}^{l+1} = \text{GTC-LN}(\mathbf{H}^{l})\in \mathbb{R}^{|\mathcal{V}'| \times d},
\end{equation}
where we employ Graph TransformerConv \cite{shi2020masked} with Layer Norm (GTC-LN) to encode \(\mathbf{H}^0\) to higher node representations \(\mathbf{H}^{l+1}\), which can capture more global structural information and lay the foundation for generating high factually correct responses. 

\begin{equation}
\mathbf{h}_{\text{mean}} = \frac{1}{|\mathcal{V}'|} \sum_{v \in \mathcal{V}'}\mathbf{H}^{l+1}_v \in \mathbb{R}^{d},
\end{equation}

\begin{equation}
\mathbf{h}_{\text{graph}} = \text{GKProj} (\mathbf{h}_{\text{mean}}) \in \mathbb{R}^{d}.
\end{equation}

Here we mean pooling the node embeddings to get a global-level representation \(\mathbf{h}_{\text{mean}}\), and further, a Graph Knowledge Projector (GKProj), implemented by a feed-forward neural network, is used to transfer the \(\mathbf{h}_{\text{mean}}\) to a graph-level representation \(\mathbf{h}_{\text{graph}}\). The projector serves as a bridge between the graph and response generation.
\begin{equation}
\mathbf{z}=\text{TextEmb}(\mathcal{U}) \in \mathbb{R}^{d}.
\end{equation}

Here we encode the utterances $\mathcal{U}$ to a text embedding \(\mathbf{z}\) by the Text Embedding Layer (TextEmb) of an LLM.

We generate the token \(\text{U}^{*}_{ti}\) by sampling the probability of \(\text{U}^{*}_{ti}\) conditioned on generated tokens \(\text{U}^{*}_{t,<i}\), \(\mathbf{h}_{\text{graph}}\), \(\mathbf{z}\) and the parameter \(\theta\), formulated as follows:

\begin{equation}
\text{U}^{*}_{ti} \sim P(\text{U}^{*}_{ti} \mid \text{U}^{*}_{t,<i}, \mathbf{h}_{\text{graph}}, \mathbf{z}; \theta).
\end{equation}

We train the model using the following cross-entropy loss:

\begin{equation}
\mathcal{L}_{\text{CE}} = - \sum_{i=1}^{T} \sum_{c=1}^{C} y_i^{(c)} \log p_i^{(c)},
\end{equation}
where \( p_i^{(c)} = P(\text{U}^{*}_{ti} = c \mid \text{U}_{t,<i}, \mathbf{h}_\text{graph}, \mathbf{z}; \theta) \) is the model-predicted probability over vocabulary class \( c \) at step \( i \). To retain the pretrained knowledge of the LLM, we adopt LoRA combined with cross-entropy to train our graph model. We refer the generation \(\text{U}^{*}_{t}\) to GA-DRG. 

\subsection{Dialogue Fact Score}
\label{Section:evaluation-fact}

We propose the Dialogue Fact Score that incorporates dialogue context and explicitly handles cases where supporting information is unavailable.

We redesign the atomic fact extraction and verification prompts used in computing FactScore. In the redesigned atomic fact extraction prompt, only complete sentences are eligible for splitting, and the model is instructed not to introduce any additional information when generating.

For fact verification, we revise the prompt to incorporate dialogue history and introduce a new class: ``Not Enough Information''. If an atomic fact is not a verifiable factual claim or lacks direct support from any known source, the model outputs ``Not Enough Information.''

We leverage the LLaMA 3.3 70B Instruct model to compute the dialogue fact score. Since the FactScore does not define the knowledge retrieval process for dialogue, we extract entities from atomic facts and dialogue as Wikipedia titles to retrieve passages as supporting knowledge.

\section{Experiment}
\subsection{Dataset}
Two public knowledge-driven and English dialogue datasets, OpendialKG \cite{moon2019opendialkg} and HybriDialogue \cite{nakamura2022hybridialogue}, are used in our work. OpenDialKG consists of a recommendation task related to movies and books and a chit-chat task related to sports and music. HybriDialogue is an open-domain and information-seeking dialogue dataset. 

\subsection{Conventional Evaluation Metrics}
We aim to evaluate factuality and quality for dialogue response generation. 

BLEU \cite{reiter2018structured}, ROUGE-L \cite{lin2004rouge}, Perplexity \cite{jelinek1977perplexity} and F1 score \cite{nan2021entity} are employed to evaluate the quality and fact consistency of the generated responses automatically. 

For human evaluation, \emph{Coherence} checks if the response makes sense and matches the previous context. \emph{Fluency} looks at whether the response is grammatically correct and sounds natural. \emph{Informativeness} measures if the response adds useful or new information to the conversation. We randomly extracted 50 samples for each model and recruited two annotators for annotation. HybriDialogue achieves Agreement/Kappa scores of 0.83/0.41 for Coherence, 0.90/0.53 for Fluency, and 0.89/0.53 for Informativeness, while OpendialKG scores 0.83/0.47, 0.96/0.27, and 0.80/0.60 on the same metrics, respectively. However, Kappa is less effective in reflecting the true agreement under imbalanced conditions. In the fluency task of OpendialKG, one class accounts for approximately 97\% of the data, making it hard for kappa to accurately measure the agreement.

\subsection{Modified Evaluation Metrics}
We adopt both the dialogue fact score and the NEIP (Not Enough Info Proportion) to evaluate factuality comprehensively. Two human annotators manually assessed the dialogue fact score with 100 random samples on two datasets, respectively, as shown in Table~\ref{table:factuality-agreement}. The Cohen's Kappa scores are both above 0.6, indicating substantial agreement between humans and the dialogue score.  

\begin{table}[ht]
\centering
\begin{tabular}{lcc}
    \toprule
    \textbf{Datasets}   & \textbf{Raw Agreement} & \textbf{Cohen's Kappa} \\
    \toprule
    HybriDialogue & 0.78 & 0.65 \\
    OpendialKG & 0.78 & 0.67 \\
    \bottomrule
\end{tabular}
\caption{Raw agreement and Cohen's Kappa between ground truth and LLM dialogue fact scores.}
\label{table:factuality-agreement}
\end{table}

\setlength{\tabcolsep}{4pt}
\begin{table*}[ht]
\centering
\begin{tabular}{l|cccccc|cccccc}
\toprule
\multicolumn{1}{c}{} 
& \multicolumn{6}{|c}{\textbf{OpendialKG}} 
& \multicolumn{6}{|c}{\textbf{HybriDialogue}} \\
\cmidrule(lr){2-7} \cmidrule(lr){8-13}
\textbf{Methods} 
& Fact* & NEIP$\downarrow$* & F1 & PPL$\downarrow$ & BLEU & ROUGE 
& Fact* & NEIP$\downarrow$* & F1 & PPL$\downarrow$ & BLEU & ROUGE \\
\midrule
ChatGLM-6B 
& 70.91 & 50.13 & 15.14 & \textbf{12.98} & \textbf{1.42} & \textbf{14.24}
& 46.58 & 58.19 & 39.57 & 23.66 & 4.62 & 24.03 \\
\quad+BM25 
& 70.99 & 49.07 & 14.92 & 13.15 & 1.38 & 14.02
& 53.32 & 55.18 & 40.26 & 23.77 & 4.68 & 24.19 \\
\quad+KAPING 
& 72.37 & 48.74 & 15.34 & 13.15 & 1.31 & 13.93
& 55.84 & 54.51 & \textbf{40.99} & \textbf{23.58} & 4.56 & 24.07 \\
\quad+\textbf{TG-DRG-NR} 
& 74.64 & 48.73 & \textbf{15.40} & 13.31 & 1.37 & 14.07
& 58.15 & 53.07 & 40.96 & 23.95 & \textbf{4.71} & \textbf{24.43} \\
\quad+\textbf{TG-DRG-R} 
& \textbf{76.46} & \textbf{37.93} & 15.33 & 14.41 & 1.34 & 13.47
& \textbf{59.10} & \textbf{50.99} & 40.92 & 26.03 & 4.53 & 24.23 \\
\midrule
Flan-T5-Large 
& 67.76 & 54.94 & \textbf{16.35} & 18.37 & 2.57 & \textbf{13.00}
& 60.04 & 50.48 & 40.04 & 30.89 & \textbf{9.61} & \textbf{27.12} \\
\quad+BM25 
& 69.32 & 56.09 & 16.25 & 18.92 & \textbf{2.65} & 12.90
& 67.42 & 47.43 & 38.50 & 32.79 & 8.71 & 25.81 \\
\quad+KAPING 
& 68.34 & 55.79 & 16.03 & \textbf{18.18} & 2.56 & 12.72
& 68.92 & 46.56 & 40.41 & 31.96 & 8.92 & 26.43 \\
\quad+\textbf{TG-DRG-NR} 
& 71.56 & 54.28 & 16.09 & 18.89 & 2.60 & 12.66
& 70.79 & 43.52 & 39.70 & \textbf{30.81} & 8.90 & 26.14 \\
\quad+\textbf{TG-DRG-R} 
& \textbf{74.25} & \textbf{28.53} & 15.00 & 20.22 & 2.19 & 11.27
& \textbf{74.41} & \textbf{33.68} & \textbf{40.63} & 34.01 & 8.20 & 24.63 \\
\midrule
Flan-T5-XXL 
& 69.29 & 36.62 & 14.07 & 23.37 & \textbf{3.44} & \textbf{13.63}
& 53.06 & 50.62 & 44.07 & 37.47 & \textbf{11.34} & 29.62 \\
\quad+BM25 
& 72.06 & 31.98 & 13.69 & 23.18 & 2.68 & 12.71
& 60.40 & 48.79 & \textbf{44.65} & 38.89 & 11.08 & \textbf{29.75} \\
\quad+KAPING 
& 73.00 & 32.07 & 13.85 & 23.22 & 2.91 & 12.82
& 63.84 & 48.94 & 44.13 & 37.88 & 10.95 & 29.44 \\
\quad+\textbf{TG-DRG-NR} 
& 76.20 & 31.42 & \textbf{14.50} & 23.07 & 3.16 & 13.06
& 66.62 & 46.49 & 43.91 & 41.35 & 10.93 & 29.30 \\
\quad+\textbf{TG-DRG-R} 
& \textbf{77.77} & \textbf{20.35} & 13.78 & \textbf{20.03} & 2.33 & 10.69
& \textbf{69.34} & \textbf{39.23} & 43.65 & \textbf{32.48} & 9.89 & 28.19 \\
\midrule
G-Retriever 
& 84.79 & 62.65 & 11.70 & 14.98 & 3.79 & \textbf{26.00}
& 70.29 & 55.77 & 42.05 & 49.66 & 20.36 & 39.53 \\
\textbf{GA-DRG} 
& \textbf{87.73} & \textbf{59.20} & \textbf{13.75} & \textbf{14.43} & \textbf{3.99} & 25.86
& \textbf{73.00} & \textbf{54.87} & \textbf{43.98} & \textbf{39.12} & \textbf{20.85} & \textbf{40.77} \\
\bottomrule
\end{tabular}
\caption{The experimental results on the OpendialKG and HybriDialogue dataset. The column heads with * indicate our primary metrics for evaluating factuality. Bold models mean generating responses with our proposed methods. The bold number is the best result within each category. NEIP denotes the ``not enough information'' proportion, representing the percentage of atomic facts that either lack direct knowledge support or fail to qualify as factual claims. PPL indicates Perplexity, and a lower PPL means more fluency. Fact, BLEU and ROUGE denote Dialogue Fact Score, BLEU-4 and ROUGE-L, respectively.}
\label{table:Main-results}
\end{table*}

\subsection{Baseline Methods}
Several widely recognised LLMs, such as ChatGLM \cite{zeng2022glm} and Flan-T5 \cite{chung2022scaling}, have been selected for this work. Furthermore, we compare our proposed TG-DRG-NR, TG-DRG-R and GA-DRG with BM25 \cite{robertson2009probabilistic}, KAPING \cite{baek2023knowledge}, G-Retriever \cite{he2024g}.

\subsection{Experimental Setup}

We provide detailed information regarding the experimental setup of our framework and the evaluation metrics used.

Except for the GNN settings where the rank and alpha parameters are set to 8 and 16, respectively, the rank and alpha parameters for LoRA \cite{hu2021lora} are consistently set to 32. As G-retriever is implemented using Llama2 7B \cite{touvron2023llama} with LoRA, we adopt the same model under identical settings to ensure fairness. We utilise the \texttt{multi-qa-mpnet-base-dot-v1} version for the baseline KAPING. All experiments were run a single time.

We utilised the \texttt{evaluate} package to compute BLEU-4 and ROUGE-L scores. Besides, the PPL is calculated using GPT-2 \cite{radford2019language}.

We used an A100 GPU for fine-tuning and inference. We utilised two A100 GPUs for the computation of the dialogue fact score. All A100 GPUs we used in this work have 80GB of memory.

\subsection{Experimental Result and Analysis}
We compare our proposed TG-DRG, which leverages a textualised graph, and GA-DRG, which utilises a GNN-based graph encoding approach, against LLM-based baselines, as well as BM25, KAPING, and G-Retriever. Table~\ref{table:Main-results} shows the experimental results of different methods on two public datasets. We analyze the generated responses from the perspectives of factuality, text quality and model scale.

\paragraph{Factuality}
With knowledge augmented methods, the dialogue fact score rises in all LLMs, and our TG-DRG-NR has a higher increase than KAPING and BM25. From TG-DRG-R, we can see a continuous improvement in dialogue fact score, which means the dialogue reformulation also contributes a slight improvement to the factuality.

The NEIP is the proportion of atomic facts that can not be directly found in external knowledge to support or which are not factual claims, and the lower one indicates the responses contain more verifiable facts. Compared to baseline LLMs, knowledge-augmented methods generally help reduce this proportion and produce more verifiable facts. From the table~\ref{table:Main-results}, we observe a consistent decrease in NEIP for the TG-DRG-R compared to KAPING and BM25. Furthermore, we noticed a marked improvement in NEIP with TG-DRG-NR. This suggests that reformulating the dialogue can substantially enhance factuality.

Notably, the F1 score is not always consistent with the fact score. For instance, the trend of the F1 score shown in Flan-T5-XXL is almost opposite to the dialogue fact score on the OpendialKG dataset. The reasons can be attributed to either the reference responses lack explicit answers or the F1 score ignores the semantic aspects, and two sentences with the same entities can express contrary meanings.

In short, knowledge-augmented methods improve the factuality of LLMs, including the dialogue fact score and NEIP. Our proposed frameworks perform better than G-Retriever, KAPING and BM25 in the above metrics.

\paragraph{Text Quality}
With the KG, the BLEU-4 and ROUGE-L scores fluctuate across two datasets. Our proposed TG-DRG-NR generally perform better than other knowledge-enhanced baselines, while we see a drop with TG-DRG-R. Our proposed GA-DRG achieves the highest BLEU and ROUGE-L scores, surpassing the SOTA G-retriever and demonstrating the superiority of our method.

There is a slight growth in PPL of smaller language models like ChatGLM and Flan-T5-Large: incorporating the KG decreases text fluency. After analysis, we found that the relatively smaller LLMs sometimes can not understand the KG well and thus generate unpolished responses.

Overall, knowledge-augmented methods show variable performance across different LLMs in terms of BLEU-4 and ROUGE-L scores, and they tend to increase PPL slightly for smaller language models. Our proposed GA-DRG gains the best performance in BLEU-4 on both datasets and ROUGE-L on HybriDialogue, indicating a higher similarity between responses and references.

\paragraph{Model Scale} We conduct experiments with different scales of LLMs. The Flan-T5 series LLMs are beneficial in analysing the correlation between size and performance. 

As model scale grows, dialogue fact score and NEIP show differing trends: factuality declines on HybriDialogue but improves on OpendialKG, which contradicts our intuition that larger models ensure consistent factuality.

\begin{table}[ht]
\centering
\begin{tabular}{lcccc}
\toprule
 & \multicolumn{2}{c}{\textbf{OpendialKG}} & \multicolumn{2}{c}{\textbf{HybriDialogue}} \\ 
\cmidrule(lr){2-3} \cmidrule(lr){4-5}
\textbf{Methods} & \textbf{Fact} & \textbf{NEIP$\downarrow$} & \textbf{Fact} & \textbf{NEIP$\downarrow$} \\ 
\midrule
w/o Know. Sel. & -1.1 & +1.2 & -1.1 & -0.11 \\
w/o Dial. Reform. & -1.46 & +0.44 & -2.3 & +0.48 \\
w/o Graph Repr. & -1.00 & -1.52 & -3.41 & -0.03 \\
\bottomrule
\end{tabular}
\caption{An ablation study on GA-DRG, where ``w/o'' denotes the removal of components: Knowledge Selection (Know. Sel.), Dialogue Reformulation (Dial. Reform.), and Graph Representation (Graph Repr.).}
\label{table:ablation_study}
\end{table}

\subsection{Ablation Study}
We conduct an ablation study with GA-DRG, shown in Table~\ref{table:ablation_study}. We remove our components one by one from GA-DRG to observe the performance. We only report our aim---factually related metrics. The results show a noticeable decline in performance after removing each component, underscoring their critical roles in maintaining the factual accuracy of GA-DRG.

\subsection{Human Evaluation Results}

\setlength{\tabcolsep}{3pt}
\begin{table}[ht]
\centering
\begin{tabular}{lcccccc}
\toprule
 & \multicolumn{3}{c}{\textbf{OpendialKG}} & \multicolumn{3}{c}{\textbf{HybriDialogue}} \\ 
\cmidrule(lr){2-4} \cmidrule(lr){5-7}
 \textbf{Methods} & \textbf{Cohe.} & \textbf{Flu.} & \textbf{Info.} & \textbf{Cohe.} & \textbf{Flu.} & \textbf{Info.} \\ 
\midrule
ChatGLM-6B       & 1.99 & 1.91 & 1.11 & 1.90 & \textbf{1.99} & 1.71 \\ 
+BM25            & 1.99 & 1.89 & 1.09 & 1.85 & 1.97 & 1.75 \\ 
+KAPING          & 1.98 & 1.95 & 1.18 & 1.84 & 1.98 & 1.71 \\ 
+\textbf{TG-DRG-NR}        & \textbf{2.00} & 1.94 & 1.11 & 1.93 & 1.98 & 1.76 \\ 
+\textbf{TG-DRG-R}         & 1.98 & \textbf{1.97} & \textbf{1.28} & \textbf{1.94} & 1.96 & \textbf{1.85} \\ 
\midrule
Flan-T5-XXL      & \textbf{1.56} & \textbf{2.00} & 1.36 & 1.66 & 1.69 & 1.74 \\ 
+BM25            & 1.38 & 1.94 & 1.49 & \textbf{1.68} & 1.71 & \textbf{1.80} \\ 
+KAPING          & 1.46 & 1.99 & 1.53 & 1.58 & 1.78 & 1.77 \\ 
+\textbf{TG-DRG-NR}        & 1.52 & 1.98 & 1.39 & 1.62 & 1.76 & 1.77 \\ 
+\textbf{TG-DRG-R}         & 1.47 & 1.97 & \textbf{1.65} & 1.55 & \textbf{1.85} & 1.76 \\ 
\midrule
G-retriever      & 1.96 & 1.96 & \textbf{0.71} & 1.92 & \textbf{1.82} & \textbf{1.84} \\ 
\textbf{GA-DRG}           & \textbf{2.00} & \textbf{2.00} & 0.60 & \textbf{1.96} & 1.81 & 1.82 \\ 
\bottomrule
\end{tabular}
\caption{Human evaluation results for Coherence (Cohe.), Fluency (Flu.), and Informativeness (Info.). The best results are marked within each category.}
\label{tab:human_evaluation}
\end{table}
We present the human evaluation results from three perspectives: coherence, fluency, and informativeness, as shown in Table~\ref{tab:human_evaluation}. Each aspect is assessed on a three-level scale: positive (2 points), neutral (1 point), or negative (0 points) (e.g., coherent, neutral, incoherent). The average scores for each aspect are then calculated based on these ratings.

Our proposed frameworks TG-DRG and GA-DRG demonstrate marked improvements in coherence across various settings, with the exception of the Flan-T5-XXL model. In terms of fluency, our methods achieve the highest scores for Flan-T5-XXL on HybriDialogue, as well as for ChatGLM and graph-based methods on OpendialKG. Overall, the proposed methods maintain strong performance in fluency and coherence. Additionally, the TG-DRG generally enhance informativeness.

\begin{table}[ht]
\centering
\begin{tabular}{p{8cm}}
\toprule
\textbf{Case 1} \\
\midrule
\textbf{Dialogue \{Original / Coreference Resolved\}:} \\
\textbf{Speaker A}: What do you think about Wladimir Klitschko? \\
\textbf{Speaker B}: Not a whole lot but I do know \{he / Wladimir Klitschko\} is married to Hayden Panettiere from the show \textit{Nashville}. \\
\textbf{Speaker A}: I have watched \{that / \textit{Nashville}\} before, is \{she / Hayden Panettiere\} in any other show? \\
\textbf{Speaker B}: \{She / Hayden Panettiere\} played a voice on the cartoon \textit{A Bug's Life}. Other than that I can't think of where I've seen \{her / Hayden Panettiere\}. \\
\textbf{Speaker A}: Ok, I didn't know \{she / Hayden Panettiere\} voiced a character in \{that / \textit{A Bug's Life}\}. \\
\midrule
\textbf{Reference}: I was surprised to hear that Kevin Spacey also voiced a character in that movie. \\
\midrule
\textbf{Flan-T5-XXL}: She voiced a character in the cartoon \textit{A Bug's Life}. \\
\textbf{BLEU-4}: 13.04; \textbf{ROUGE-L}: 30.8 \\
\midrule
\textbf{Key Knowledge}: (Dot, voice actor, Hayden Panettiere) \\
\textbf{TG-DRG-R}: Hayden Panettiere voiced the character Dot in \textit{A Bug's Life}. \\
\textbf{BLEU-4}: 0; \textbf{ROUGE-L}: 23.1 \\
\bottomrule
\end{tabular}
\caption{Dialogue and LLMs' responses related to Hayden Panettiere, before and after coreference resolution.}
\label{table:Case-Hayden-Panettiere}
\end{table}

\subsection{Case Study}
\label{Section:Case-Study}


We use a case about Hayden Panettiere shown in Table~\ref{table:Case-Hayden-Panettiere} to illustrate our work. From the case, the dialogue contains pronouns, e.g. ``she'', challenging the generation of factually correct responses. We can see the response from Flan-T5-XXL is repeating the previous utterance ``She played a voice on the cartoon A Bug’s Life...''. After dialogue reformulation and dialogue sense knowledge selection, our TG-DRG-R is able to generate a more factual and detailed response. However, from Table~\ref{table:Main-results}, we observe that TG-DRG-R's response achieves lower BLEU-4 and ROUGE-L scores compared to those of Flan-T5-XXL. This suggests that BLEU-4 and ROUGE-L may not effectively capture the quality of responses in open-domain dialogue tasks.



\section{Conclusion}
Our work addresses the limitations of previous methods in handling factual dialogue response generation, particularly issues arising from complex coreference in the dialogue. To this end, we propose two novel frameworks, TG-DRG and GA-DRG, which feature reasoning-guided dialogue reformulation and dialogue sense knowledge selection, leveraging graph-based generation techniques. The results demonstrate that these frameworks noticeably enhance the factual accuracy of dialogue responses compared to baseline methods, including the SOTA G-retriever.

Additionally, we propose the dialogue fact score as a novel metric to evaluate the factuality of dialogue responses. This metric shows high consistency with human evaluations and proves to be more reliable than traditional metrics such as BLEU-4, ROUGE-L, and F1, which often fail to capture factual consistency in open-domain dialogue tasks.

While our approach shows noticeable improvements, certain limitations remain. For instance, the validity of triples from external knowledge graphs may change over time, potentially impacting factuality. Addressing this challenge will be a focus of our future work.

\bibliography{aaai2026}

%

\clearpage
\appendix
\section{Prompts}
\label{sec:prompts}
In this section, we show the prompts used in our work. 

The prompt for reasoning-guided dialogue reformulation is described in Table~\ref{table:dialogue-reformulation-prompt}. The prompts for the knowledge-augmented dialogue response generation, used in TG-DRG, and atomic fact splitting are shown in Table~\ref{table:DRG-Atomic-prompt}. The prompt for fact verification of dialogue fact score prompts is described in Table~\ref{table:dialogue-fact-score-prompt}.

\begin{table*}
  \centering
   \begin{tabular}{l}
    \toprule
    \textbf{Dialogue Reformulation Prompt}\\
    \toprule
    \hline
    {\parbox[t]{17cm}{
  You are tasked with resolving all pronouns and references in the given dialogue to their explicit entities. Use CoT (Chain of Thought) reasoning to identify what each pronoun or reference corresponds to. Do not answer any questions; your only goal is to perform co-reference resolution.
  
  Instructions:
  
  1. Analyze the dialogue and process each turn in the conversation.
  
  2. For every pronoun, ambiguous term, or reference, trace back in the conversation to determine its explicit entity or subject.
  
  3. Clearly document your CoT reasoning for each resolution.
  
  4. Provide the explicit reference for each pronoun or ambiguous term.
  
  Output Format:
  
  **Chain of Thought**: [Your reasoning process for resolving the references]
  
  **Resolved Dialogue**: [The dialogue with all pronouns and references resolved] \\
     \textbf{Dialogue}: \{Dialogue\}
    }}
    \\
    \bottomrule
   \end{tabular}
   \caption{The prompt for reasoning-guided dialogue reformulation.}
\label{table:dialogue-reformulation-prompt}
\end{table*}

\begin{table*}[ht]
\centering
\begin{tabular}{ll}
\toprule
\textbf{Dialogue Response Generation Prompt}   & \textbf{Atomic Fact Splitting Prompt} \\ 
\toprule 
\toprule
\parbox[t]{8cm}{
\textbf{Knowledge}: \{Selected Triples\}\\
\textbf{Dialogue}: \{Dialogue Context\} \\
Given the above knowledge and dialogue, please respond to the input below and ensure the response is fluent and fact-consistent in English.\\
\textbf{Input}: \{Last Utterance\} \\
\textbf{Response}: ... \\
} & \parbox[t]{9cm}{
\textbf{\{Examples..}\} \\
If the following input is an incomplete sentence or a phrase, please output it exactly as it is. \\
Otherwise, if it is a complete sentence, split it into atomic sentences based only on the given information, without adding any additional information or making inferences: \\
\textbf{Input}: \{Response\} \\
\textbf{Output} ...:
}
\\
\bottomrule
\end{tabular}
\caption{The prompts for dialogue response generation (left) and atomic fact splitting (right).}
\label{table:DRG-Atomic-prompt}
\end{table*}

\begin{table*}[ht]
\centering
\begin{tabular}{l}
\toprule
\textbf{Dialogue Fact Score Prompt}   \\ \toprule \toprule
{\parbox[t]{17cm}{
Instruction:\\
The statement is part of a response in a dialogue. Evaluate the statement strictly based on the provided knowledge source and dialogue history only.\\
If the statement is not a factual claim (e.g., opinion, question, or unclear assertion), output: "no enough information."\\

If it is a factual claim:\\
Output true if the statement is directly supported by evidence in the knowledge source or dialogue history.\\
Output false if the statement is directly contradicted by the knowledge source or dialogue history.\\
Output no enough information if there is no direct evidence for or against the statement.\\

Important:\\
Do not use your intern knowledge or make inferences.\\
Please only output your final answer and do not output any explanations.

\textbf{Evidence}: \{Wikipedia Passages\} \\
\textbf{Dialogue history}: \{Dialogue\} \\
\textbf{Speaker A}: \{Speaker\} \\
\textbf{Statement}: \{Atomic Fact\}
}} \\
\bottomrule
\end{tabular}
\caption{The prompt for fact verification of dialogue fact score.}
\label{table:dialogue-fact-score-prompt}
\end{table*}

\section{Datasets Details}
\label{sec:datasets}
This section provides a detailed description of the datasets, listed as follows:

\paragraph{OpenDialKG} It is a knowledge-driven dialogue dataset. The dataset contains 13,802 samples. After processing, some samples are not valid in the task, so we remove them. Finally, we randomly selected 15\% (1,962 samples) for the validation set, another 15\% (1,973 samples) for the test set, and the remaining 70\% (9,120 samples) for the training set. The dataset provides a number of triples extracted from Freebase \cite{bast2014easy}. These triples were collected several years ago. To address any potential issues with outdated information, we re-extract the triples from Wikidata.  OpendialKG is released under CC-BY-NC-4.0 license, which permits non-commercial research use.

\paragraph{HybriDialogue} It is an open-domain dialogue dataset. It is constructed by splitting complex questions into multi-turn dialogue. The original dataset does not offer triples, so we collect them by matching the entities with triples from Wikidata. The training set contains 4,359 samples, while the validation set includes 242 samples, and the test set consists of 243 samples. HybriDial follows the MIT license \cite{nakamura2022hybridialogue}, which allows both commercial and non-commercial use.

\begin{table*}[ht]
\centering
\begin{tabular}{p{17cm}}
\toprule
\textbf{OpendialKG Example} \\ \toprule \toprule
\textbf{Last Utterance}: Could you recommend some movies staring Chiaki Kuriyama?  \\
\textbf{Selected triples from BM25}: (Chiaki Kuriyama, given name, Chiaki) (Chiaki Kuriyama, family name, Kuriyama) (Chiaki Kuriyama, place of birth, Tsuchiura) (Chiaki Kuriyama, occupation, film actor) (Chiaki Kuriyama, instrument, voice) ... \\
\textbf{Selected triples from KAPING}: (Chiaki Kuriyama, given name, Chiaki) (Chiaki Kuriyama, occupation, model) (Chiaki Kuriyama, occupation, singer) (Chiaki Kuriyama, occupation, actor) (Chiaki Kuriyama, occupation, fashion model) ... \\ 
\textbf{Selected triples from our method}: (Kill Bill Volume 1, cast member, Chiaki Kuriyama), (Into the Sun, cast member, Chiaki Kuriyama), (Kagen no Tsuki, cast member, Chiaki Kuriyama), (Kamogawa Horumo, cast member, Chiaki Kuriyama), (Gonin, cast member, Chiaki Kuriyama) ... \\ \bottomrule
\textbf{HybriDialogue Example} \\ \toprule \toprule 
\textbf{Last Utterance}: Hi. Can you tell me who Judd Trump is? \\
\textbf{Selected triples from BM25}: (Judd Trump, given name, Judd)(Judd Trump, family name, Trump)(2021 Champion of Champions, winner, Judd Trump)(Judd Trump, victory, 2016 China Open)(Judd Trump, victory, 2011 China Open) ...\\
\textbf{Selected triples from KAPING}: (Judd Trump, occupation, snooker player) (Judd Trump, nickname, The Ace) (Judd Trump, given name, Judd) (Judd Trump, family name, Trump) (Judd Trump, place of birth, Whitchurch) ... \\
\textbf{Selected triples from our method}: (2019 World Snooker Championship, winner, Judd Trump), (Judd Trump, award received, Snooker Hall of Fame), (Judd Trump, country of citizenship, United Kingdom), (Judd Trump, sport, snooker), (Judd Trump, award received, player of the year award) ...\\ \bottomrule
\end{tabular}
\caption{The examples extracted from OpendialKG and HybriDialogue dataset. The triples are retrieved from Wikidata and selected using different methods.}
\label{table:dataset-example}
\end{table*}

Table~\ref{table:dataset-example} presents the two examples of OpendialKG and HybridDial datasets with selected triples from different methods.

\section{Examples generated from GPT-4o}
\label{Section:GPT4o-examples}
We employ GPT-4o to generate data for dialogue sense knowledge selection and dialogue reformulation from the training dataset, where we collected 29K samples and 35.4K for them. The examples for fine-tuning are listed in Tables~\ref{table:relevance} and ~\ref{table:cot_coreference_resolution}. 

\begin{table*}[h]
    \centering
    \begin{tabular}{ p{7.5cm}  p{7.5cm}  c }
        \toprule
        \textbf{Last Utterance} & \textbf{Triple} & \textbf{Relevance} \\
        \toprule
        Which teams has the Colorado Avalanche played against in the playoffs? & (Colorado Avalanche, sport, ice hockey) & Irrelevant \\
        Hi! Do you know anything about Lutheran churches? & (Lutheranism, described by source, Brockhaus and Efron Encyclopedic Dictionary) & Irrelevant \\
        Tell me more about Upper Canada District School Board? & (Upper Canada District School Board, headquarters location, Brockville) & Relevant \\
        Hi, do you by any chance know anything about Vladimir Horowitz? & (Stefanos Korkolis, student of, Vladimir Horowitz) & Relevant \\
        \bottomrule
    \end{tabular}
    \caption{The dialogue and corresponding triple are taken from the training dataset and Wikidata, respectively, with only the last utterance of the dialogue used here.. The relevance is generated by GPT-4o. These examples are adopted for fine-tuning our dialogue sense knowledge selection.}
    \label{table:relevance}
\end{table*}

\begin{table*}[h]
    \centering
    \begin{tabular}{p{3.5cm}p{14cm}}
\toprule
\textbf{Section} & \textbf{Content} \\ \toprule
Original Dialogue & 
\textbf{Speaker A:} I like Lev Grossman books can you recommend some of his book

\textbf{Speaker B:} Sure! He wrote The Magician's Land and The Magicians.

\textbf{Speaker A:} Oh, I think I've read The Magicians it a High Fantasy book. \\
\midrule
CoT & 
Speaker A mentions "Lev Grossman" as the author they like. This establishes that "his" in their request for book recommendations refers to Lev Grossman.

Speaker B mentions "He wrote The Magician's Land and The Magicians." The pronoun "He" refers to Lev Grossman, as he is the author being discussed.

Speaker A uses "it" in "I think I've read The Magicians it a High Fantasy book." The pronoun "it" refers to "The Magicians," as it is the last explicitly mentioned book in the conversation. \\
\midrule
Reformulated Dialogue & 
\textbf{Speaker A:} I like Lev Grossman books can you recommend some of Lev Grossman's books

\textbf{Speaker B:} Sure! Lev Grossman wrote The Magician's Land and The Magicians.

\textbf{Speaker A:} Oh, I think I've read The Magicians. The Magicians is a High Fantasy book. \\
\bottomrule
\end{tabular}
\caption{An example generated by GPT-4o illustrates how CoT reasoning works in dialogue reformulation.}
\label{table:cot_coreference_resolution}
\end{table*}

\begin{table*}[ht]
\centering
\begin{tabular}{p{\textwidth}}
\toprule
\textbf{Example 1} \\
\midrule
\textbf{Dialogue:} \\
\textbf{Speaker A}: I enjoy Charlaine Harris's work, can you name something similar? \\
\textbf{Speaker B}: \textit{The Southern Vampire Mysteries} I would recommend for you. \\
\textbf{Speaker A}: When was it released? \\
\textbf{Speaker B}: 2004. \\
\midrule
\textbf{Speaker A / Reformulated} \\
How many books are in the series? / How many books are in \textbf{The Southern Vampire Mysteries} series? \\

\midrule
\textbf{Label}: Good \\
\textbf{Reason}: The reference ``The series'' in the last utterances has been resolved as ``The Southern Vampire Mysteries'' correctly. \\
\bottomrule
\toprule
\textbf{Example 2} \\
\midrule
\textbf{Speaker A}: Could you recommend a book written by author Fannie Flagg? \\
\textbf{Speaker B}: \textit{Welcome to the World, Baby Girl} and \textit{I Still Dream About You} are great books that I would suggest. You read them? \\
\textbf{Speaker A}: No, I have read neither of those. What are their genres? \\
\textbf{Speaker B}: It is adventure and is also a movie released in 2010, I would suggest reading the book then watching the movie. \\
\midrule
\textbf{Speaker A / Reformulated:} \\
Yes, I will definitely read the book first. Can you add it to my list? / Yes, I will definitely read \textbf{``Welcome to the World, Baby Girl''} first. Can you add \textbf{``Welcome to the World, Baby Girl''} to my list?\\
\midrule
\textbf{Label}: Neutral \\
\textbf{Reason}: The reference ``the book'' in the dialogue is ambiguous. So the reformulated utterance may not be correct.\\
\bottomrule
\toprule
\textbf{Example 3} \\
\midrule
\textbf{Dialogue:} \\
\textbf{Speaker A}: Cool! do you know the occasion name of the Video event? \\
\textbf{Speaker B}: Yes, it is NeXT's internal marketing strategy video. \\
\textbf{Speaker A}: Great! do you know the city where that video took place? \\
\textbf{Speaker B}: Yes, it was Redwood City. \\
\midrule
\textbf{Speaker A / Reformulated:} \\
Good! do you know how many miles \textbf{that city} far from south of San Francisco? / Good! do you know how many miles \textbf{that city} far from south of San Francisco? \\
\midrule
\textbf{Label}: Bad \\
\textbf{Reason}: The reference ``that city'' should be resolved, but the model fails to resolve it.\\
\bottomrule
\end{tabular}
\caption{The manual annotated examples for dialogue reformulation. In human evaluation, we only evaluate the last utterance.}
\label{table:dialogue_reformulation_examples}
\end{table*}

\section{Conventional Evaluation Metrics in Detail}
We list all the details of conventional evaluation metrics as follows:
\label{Section:conventional-evaluation-metrics}
\begin{itemize}\itemsep=0pt
\item \textbf{BLEU} \cite{reiter2018structured}, used in the machine translation initially, calculates the n-gram overlap between generated text and references and reflects the correlation between generated text and human writing. 

\item \textbf{ROUGE-L} \cite{lin2004rouge} is calculated based on the length of the longest common subsequence, which captures the word order and sentence-level structure from the ground truth. 

\item \textbf{PPL} \cite{jelinek1977perplexity} is widely used in measuring text fluency of generated output, in which calculation is based on the language model. 

\item \textbf{F1 Score} \cite{nan2021entity} is based on the precision and recall of the extracted entities matching between the generated response and ground truth. 

\end{itemize}

\begin{table*}[ht]
\centering
\begin{tabular}{p{\textwidth}}
\toprule
\textbf{Case 2} \\ 
\toprule
\textbf{Original Dialogue:} \\
\textbf{Speaker A}: Hello! do you know fictional marsupials? \\
\textbf{Speaker B}: Yes, it is subsidiary to the list of fictional animals and is a collection of various notable marsupial characters that appear in various works of fiction. \\
\textbf{Speaker A}: Great! can you list some literature names of fictional marsupials? \\
\textbf{Speaker B}: Yes, there are many books available including \textit{Ginger Kangaroo}, \textit{Kidding Kangaroo}, etc. \\
\textbf{Speaker A}: Good! do you know the species name of the literature \textit{Blinky Bill}? \\
\textbf{Speaker B}: Yes, it is Koala. \\
\textbf{Speaker A}: Great! do you know the author of that book? \\
\midrule
\textbf{Flan-T5-XXL:} Yes, it is written by E. Nesbit.  \\
\textbf{Fact Verification:} False\\
\midrule
\textbf{Reformulated Dialogue:} \\
\textbf{Speaker A}: Hello! do you know fictional marsupials? \\
\textbf{Speaker B}: Yes, \textbf{fictional marsupials} are subsidiary to the list of fictional animals and is a collection of various notable marsupial characters that appear in various works of fiction. \\
\textbf{Speaker A}: Great! can you list some literature names of fictional marsupials? \\
\textbf{Speaker B}: Yes, there are many books available including \textit{Ginger Kangaroo}, \textit{Kidding Kangaroo}, etc. \\
\textbf{Speaker A}: Good! do you know the species name of the literature \textit{Blinky Bill}? \\
\textbf{Speaker B}: Yes, the \textbf{species name of \textit{Blinky Bill}} is Koala. \\
\textbf{Speaker A}: Great! do you know the author of 
\textbf{\textit{Blinky Bill}}? \\
\midrule
\textbf{Key Knowledge:} (\textit{Blinky Bill}, illustrator, Dorothy Wall), (\textit{Blinky Bill}, author, Dorothy Wall) ... \\
\textbf{TG-DRG-R:} Yes, the author of \textit{Blinky Bill} is Dorothy Wall.\\
\textbf{Fact Verification:} True\\
\bottomrule
\toprule
\textbf{Case 3}      \\ \toprule
\textbf{Context}:\\
\textbf{Speaker A}: What is the capital and largest city in Uruguay?  \\
\textbf{Speaker B}: Montevideo is the capital and largest city in Uruguay. \\
\textbf{Speaker A}: What was the population of Montevideo, Uruguay, in 2011? \\
\midrule
\textbf{Flan-T5-Large}: The population of Montevideo, Uruguay, in 2011 was 59,027. \\
\midrule
\textbf{Key Knowledge}: (Montevideo, population, 309331) \\
\textbf{TG-DRG-NR}: Montevideo, population, 309331 \\
\bottomrule
\end{tabular}
\caption{The responses of cases were generated by Flan-T5 series models. In case 2, the resolved coreference is bold in the reformulated dialogue. }
\label{table:Case-Studies}
\end{table*}

\section{Details of Manual Annotation}
\label{section:factuality-agreement}
We invited several annotators to annotate the samples. All of them are well-educated and possess good English proficiency. The annotators were told that the annotated data would be used for research, and we got consent for data use.

To assess the accuracy of the samples used in fine-tuning dialogue sense knowledge selection generated from GPT-4o, we invite three annotators to this task. We randomly selected 100 samples and instructed participants to determine whether the last utterance of the dialogue is relevant to the triple. The output is either \emph{relevant} or \emph{irrelevant}. In the beginning, two annotators have 0.83 in agreement and 0.66 in Cohen's Kappa. We introduced another annotator to mediate the disagreement. We utilized the final annotated result to assess the accuracy of query-triple pairs, which is 0.81 in agreement and 0.62 in Cohen's Kappa. These results indicate a high level of accuracy.

Three annotators were involved in the assessment of the dialogue fact score task. We used the dialogue fact score prompt, described in Table~\ref{table:dialogue-fact-score-prompt}, as an instruction. Two annotators independently assessed the dialogue fact score for 100 samples from the HybriDial and OpendialKG datasets, respectively. Initially, the raw agreement scores were 0.76 for HybriDial and 0.81 for OpenDialKG. Their inter-annotator agreement, measured by Cohen's Kappa, was 0.613 for OpendialKG and 0.705 for HybriDial. To resolve discrepancies, we introduced a third annotator to mediate disagreements. Finally, we report the final agreement and Cohen's Kappa score with the evaluation model, as shown in Table~\ref{table:factuality-agreement}. 






We randomly selected 100 samples to evaluate the quality of dialogue reformulation, with two annotators involved in the process. The evaluation focuses solely on the last utterance, as it is the most crucial part of the dialogue. Each sample is manually rated as good, neutral, or bad. On the HybriDialogue dataset, the annotators achieved a raw agreement of 0.91 and a Cohen's kappa of 0.580. On the OpenDialKG dataset, they reached a raw agreement of 0.94 and a Cohen's kappa of 0.548, indicating moderate agreement. The annotation examples of dialogue reformulation are shown in Table~\ref{table:dialogue_reformulation_examples}.

\section{Baseline Descriptions}
\label{section:baseline-llms}
LLMs used in this work are presented as follows:
\begin{itemize}\itemsep=0pt
  \item \textbf{ChatGLM} \cite{zeng2022glm} is an open-source and bilingual language model, based on GLM \cite{du2021glm} architecture. The technique of ChatGLM-6B is similar to ChatGPT, optimized in dialogue tasks.
  \item \textbf{Flan-T5} \cite{chung2022scaling} proposed several instruction-based LLMs, scaling the number of tasks and model size and fine-tuning in the chain-of-thought data. We adopt Flan-T5-XXl (11B), Flan-T5-Large (780M) and Flan-T5-Small (80M) as baselines in our work.
\end{itemize}

The baseline methods are described as follows:

\begin{itemize}\itemsep=0pt

\item \textbf{No Knowledge}: We feed the prompt into LLMs to generate dialogue responses without external knowledge. 

\item \textbf{BM25} \cite{robertson2009probabilistic}: It is the ranking function between a query and several documents widely used in information retrieval. In our experiments, we also apply RefinED to retrieve triples for BM25 and replace documents with triples for retrieval. 

\item \textbf{KAPING} \cite{baek2023knowledge}: It is a popular knowledge-augmented method for question answering. Since the authors have not released the official code, we follow the experimental setup of the KAPING method: we first extract entities from the query using REfinED, rank the query with triples using MPNet, select the top-N most relevant triples, and supplement them as prompts to generate dialogue responses.

\item \textbf{G-Retriever} \cite{he2024g}: It is a retrieval-augmented generation method for textual graphs that selects informative subgraphs via a Prize-Collecting Steiner Tree, encodes them with graph neural networks, and generates responses with high accuracy, outperforming strong baselines and reducing hallucinations.




\end{itemize}

\section{Dialogue Reformulation Evaluation}
We evaluate the dialogue reformulation performance through human judgment, with annotation details provided in ~\ref{section:factuality-agreement}. Each instance is rated as good (1), neutral (0.5), or bad (0), and the final score is computed as the average. Our method achieves a score of 0.9325 on the HybriDialogue dataset and 0.9475 on the OpenDialKG dataset, demonstrating the strong effectiveness of our proposed reasoning-guided dialogue reformulation approach.

\section{Addtional Case Studies}
\label{section:appendix-case-study}
The additional case studies are shown in Table~\ref{table:Case-Studies}. 

Case 2 presents a dialogue about the book \textit{Blinky Bill}. The original dialogue contains coreference issues, making the dialogue difficult to follow and resulting in a factually incorrect response. By applying our TGDRG-R approach, which combines step-by-step reference resolution in reasoning-guided dialogue reformulation with dialogue sense knowledge selection, we achieve the generation of correct and coherent responses.

The topic in Case 3 is about the population of Montevideo, and our methods do help select the relevant triples. However, the TG-DRG-NR then generates one of the triples directly, which is not fluent and has a higher PPL i.e.\ lower \emph{quality}.

\end{document}